\documentclass[final]{cvpr}

\usepackage{times}
\usepackage{epsfig}
\usepackage{graphicx}
\usepackage{amsmath}
\usepackage{amssymb}

\usepackage{bm}
\usepackage{multirow}
\usepackage[justification=centering]{subcaption}
\usepackage{nicefrac}
\usepackage{comment}
\usepackage{enumitem}
\usepackage{adjustbox}
\usepackage{mathabx}
\usepackage{tabu}
\usepackage{booktabs}
\usepackage{diagbox}
\usepackage{cuted}
\usepackage{capt-of}
\usepackage[table,x11names]{xcolor}
\usepackage{makecell}
\usepackage{bm}
\usepackage{amsmath}

\usepackage[pagebackref=true,breaklinks=true,colorlinks,bookmarks=false]{hyperref}

\def\cvprPaperID{2907} 
\def\confYear{CVPR 2021}

\begin{document}

\title{Generative Zero-shot Network Quantization}

\author{Xiangyu He\qquad Qinghao Hu\qquad Peisong Wang\qquad Jian Cheng\\
NLPR, CASIA\\
{\tt\small xiangyu.he@nlpr.ia.ac.cn}
}
\maketitle

\begin{abstract}
Convolutional neural networks are able to learn realistic image priors from numerous training samples in low-level image generation and restoration \cite{UlyanovVL18}. We show that, for high-level image recognition tasks, we can further reconstruct ``realistic'' images of each category by leveraging intrinsic Batch Normalization (BN) statistics without any training data. Inspired by the popular VAE/GAN methods, we regard the zero-shot optimization process of synthetic images as generative modeling to match the distribution of BN statistics. The generated images serve as a calibration set for the following zero-shot network quantizations. Our method meets the needs for quantizing models based on sensitive information, \textit{e.g.,} due to privacy concerns, no data is available. Extensive experiments on benchmark datasets show that, with the help of generated data, our approach consistently outperforms existing data-free quantization methods.
\end{abstract}

\section{Introduction}
Deep convolutional neural networks have achieved great success in several computer vision tasks \cite{HeZRS16, RenHGS15}, however, we still understand little why it performs well. There are plenty of pioneering works aim at peeking inside these networks. Feature visualization \cite{olah2017feature, DeepDream, olah2020zoom} is a group of mainstream methods that enhance an input image from noise to elicit a particular interpretation. The generated images illustrate how neural networks build up their understanding of images \cite{olah2017feature}, as a byproduct, opens a path towards data-free model compression \cite{DeepInversion}.

Network quantization is an efficient and effective way to compress deep neural networks with a small memory footprint and low latency. It is a common practice to introduce a calibration set to quantize activations. To recover the degraded accuracy, training-aware quantization even requires re-training on the labeled dataset. However, in real-world applications, the original (labeled) data is commonly not available due to privacy and security concerns. In this case, zero-shot/data-free quantization becomes indispensable. The following question is \textit{how to sample data $x$ from the finite dataset $\mathcal{X}$ in the absence of original training data}.

It is intuitive to introduce noise $u\sim\mathcal{N}(0,1)$ as the input data to estimate the distributions of intermediate layers. Unfortunately, since the Single-Gaussian assumption can be too simple, the results for low-bit activation quantization are far from satisfactory \cite{GDFQ}. Due to the zero-shot setting, it is also hard to apply the well-developed GANs to image synthesis without learning the image prior to the original dataset. Very recently, a large body of works suggests that the running mean $\mu$ and variance $\sigma^2$ in the Batch Normalization layer have captured the prior distribution of the data \cite{ZeroQ, KnowledgeWithin, DFNQ, GDFQ}. The square loss on $\mu$ and $\sigma^2$ (details in Equation (\ref{eq:square_loss}) and (\ref{eq:kl_mu_var})) coupled with cross-entropy loss achieves the empirical success in zero-shot network quantization. Though the performance has been further improved over early works \cite{dreamdis, MahendranV15, olah2017feature, olah2020zoom} such as DeepDream \cite{DeepDream}, it remains unclear why these learning targets should lead to meaningful data after training. Therefore, instead of directly presenting several loss functions, we hope to better describe the training process via generative modeling, which might provide another insight into the zero-shot network quantization.

In this work, we consider the generative modeling that deals with the distributions of mean and variance in Batch Normalization layers \cite{BN}. That is, for some random data augmentations, the mean $\mu$ and variance $\sigma^2$ generated by synthesized images $I$ should look like their counterparts extracted from real images, with high probability. Recently developed generative models like GAN commonly captures $p(\cdot)$ instead of knowing one image, but we regard synthesized images $I$ as model parameters optimized in zero-shot learning. The input transformations introduce the randomness to allow the sampling on $\mu$ and $\sigma^2$. Besides, our method presents a potential interpretation for the popular Batch Normalization matching loss \cite{ZeroQ, KnowledgeWithin, DFNQ, GDFQ, DeepInversion}. Due to the insufficient sampling in each iteration, we further propose a prior regularization term to narrow the parameter space of $I$. Since the accuracy drop of low-bit activation quantization heavily relies on the image quality of the calibration set, we conduct extensive experiments on the zero-shot network quantization task. The 4-bit networks including weights and activations show consistent improvements over baseline methods on both CIFAR and ImageNet. Hopefully, this work may not be restricted to image synthesis, but shed some light on the interpretability of CNN through generating interpretable input images that agree with deep networks' internal knowledge.

\section{Related Work}
Model quantization has been one of the workhorses in industrial applications which enjoys low memory footprint and inference latency. Due to its great practical value, low-bit quantization becomes popular in recent literature.

\noindent\textbf{Training-aware quantization}. Previous works \cite{morgan1991experimental, GuptaAGN15, HubaraCSEB17, DoReFa} mainly focus on recovering the accuracy of quantized model via backward propagations, $i.e.$, label-based fine-tuning. Since the training process is similar to its floating point counterpart, the following works further prove that low-bit networks can still achieve comparable performances with full-precision networks by training from scratch \cite{Dettmers15, INQ, LouizosRBGW19, JacobKCZTHAK18}. Ultra low-bit networks such as binary \cite{Grossman89, BC, BNN, XNOR} and ternary \cite{AbramsonSM93, ChiuehG88, TWN, TTN}, benefiting from bitwise operations to replace the computing-intensive MACs, is another group of methods. Leading schemes have reached less than five points accuracy drop on ImageNet \cite{ABC, ProxyBNN, MartinezYBT20, KimK0K20}. While training-aware methods achieve good performance, they suffer from the inevitable re-training with enormous labeled data.

\noindent\textbf{Label-free quantization}. Since a sufficiently large open training dataset is inaccessible for many real-world applications, such as medical diagnosis \cite{Oncology}, drug discovery, and toxicology \cite{DrugDes}, it is imperative to avoid retraining or require no training data. Label-free methods take a step forward by only relying on limited unlabeled data \cite{learningCompression, UporDown, BannerNS19, OCS, ACIQ, OMSE}. Most works share the idea of minimizing the quantization error or matching the distribution of full precision weights via quantized parameters. \cite{BannerNS19, FQB, wang2020unsupervised, UporDown} observe that the changes of layer output play a core role in accuracy drop. By introducing the ``bias correction'' technique (minimize the differences between $\mathbb{E}(\bm{W}x)$ and $\mathbb{E}(\widehat{\bm{W}}x)$), the performance of quantized model can be further improved.

\noindent\textbf{Zero-shot quantization}. Recent works show that deep neural networks pre-trained on classification tasks can learn the prior knowledge of the underlying data distribution \cite{UlyanovVL18}. The statistics of intermediate features, \textit{i.e.,} ``metadata'', are assumed to be provided and help to discriminate samples from the same image domain as the original dataset \cite{dreamdis, DFKD}. To circumvent the need for extra ``metadata'', \cite{NayakMSRC19} treats the weights of the last fully-connected layer as the class templates then exploits the class similarities learned by the teacher network via Dirichlet sampling. Another group of methods focus on the stored running statistics of the Batch Normalization layer \cite{BN}. \cite{DeepInversion, KnowledgeWithin} produce synthetic images without generators by directly optimizing the input images through backward propagations. Given any input noises, \cite{DAFL, MicaelliS19, YooCKK19, DFNQ, MAZE, ZeroQ} introduces a generator network $\bm{g}_\theta$ that yields synthetic images to perform Knowledge Distillation \cite{HintonVD15} between teacher and student networks.

Since we have no access to the original training dataset in zero-shot learning, it is hard to find the optimal generator $\bm{g}_\theta$. Generator-based methods have to optimize $\bm{g}_\theta$ indirectly via KL divergence between categorical probability distributions instead of $\max\ \mathbb{E}_{x\sim\hat{p}}[\ln p(x)]$ in VAE/GANs. In light of this, we formulate the optimization of synthetic images $I$ as generative modeling that maximizes $\mathbb{E}[\ln p(\mu,\sigma^2; I)]$.

\begin{figure}[t]
\centering
\includegraphics[width=0.4\textwidth]{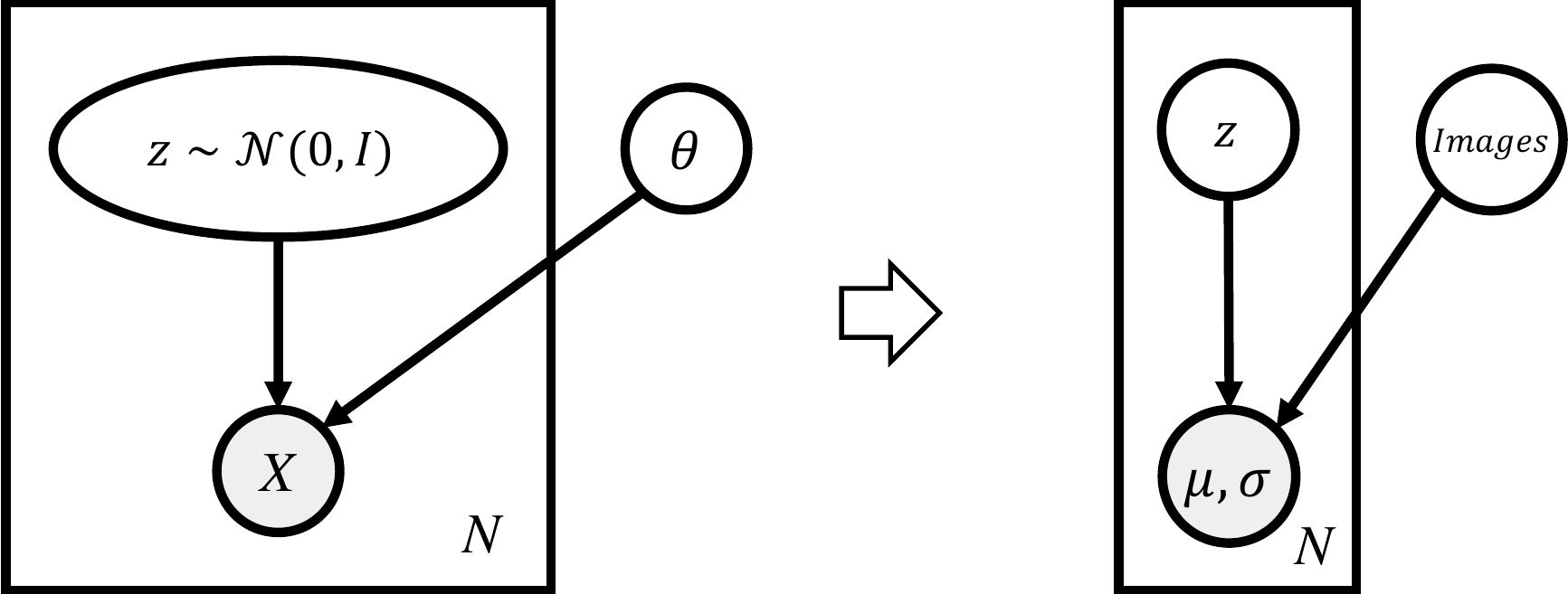}
\caption{The standard VAE represented as a graphical model (left subfigure), \textit{i.e.}, sampling from $z$ $N$ times to generate something similar to $X$ with fixed parameters $\theta$ \cite{VAE}. In this work, we regard the synthetic sampes/images $I$ as model parameters $\theta$ to be optimized during the training process. We have a family of deterministic functions $f_I(z)$ parameterized by $I$. Since $z$ is random, $f_I(z)$ is a random variable \protect\footnotemark. We hope to optimize $I$ such that $\forall z_i$ from $p(z)$, $f_I(z_i)$ can cause the pre-trained network to generate $\mu,\sigma^2$ and, with high probability, these random variables will be like the Batch-Normalization statistics \cite{BN} generated by real images.}
\label{fig:VAE}
\end{figure}
\footnotetext{Given $z_i$, we have a deterministic function $f_I(z_i)$ which applies flipping/jitter/shift to the synthetic images $I$ according to $z_i$. We can sample $z_i$ from probability density function $p(z)$ $N$ times to allow the backprop.}

\section{Approach}
\subsection{Preliminary}
A generative modeling whose purpose is to map random variables to samples and generates samples distributed according to $\hat{p}(x)$, defined over datapoints $\mathcal{X}$. We are interested in estimating $\hat{p}$ using maximum likelihood estimation to find best $p(x)$ that approximates $\hat{p}$ measured by Kullback-Leibler divergence,
\begin{equation}
\mathcal{L}(x;\theta)=\text{KL}(\hat{p}(x)||p(x;\theta))=\mathbb{E}_{\hat{p}(x)}[\ln p(x;\theta)].
\label{KL_loss}
\end{equation}
Here we use a parametric model for distribution $p$. We hope to optimize $\theta$ such that we can maximize the log likelihood.

Ideally, $p(x;\theta)$ should be sufficiently expressive and flexible to describe the true distribution $\hat{p}$. To this end, we introduce a latent variable $z$,
\begin{equation}
p(x;\theta)=\int p(x|z;\theta)p(z)dz=\mathbb{E}_{p(z)}[p(x|z;\theta)],
\end{equation}
so that the marginal distribution $p$ computed by the product of diverse distributions (\textit{i.e.,} joint distribution) can better approximate $\hat{p}$.

\begin{figure*}[t]
\centering
\includegraphics[width=0.9\textwidth]{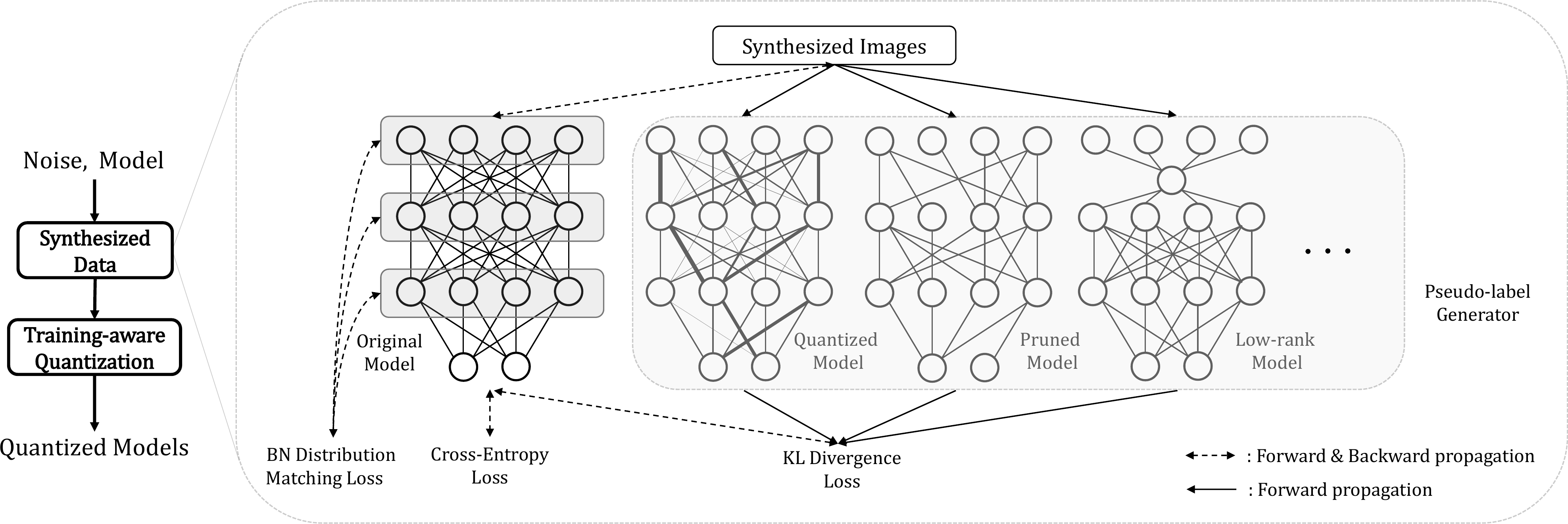}
\caption{Generative Zero-shot Network Quantization (GZNQ) setup: GZNQ uses a generative model to produce the mean and variance in Batch-Normalization (BN) layer \cite{BN}, meanwhile, optimizes the synthesized images to perform the following training-aware quantization. The pseudo-label generator consists of several data-free post-training compressed models, which serve as a multi-model ensemble or voting classifier.}
\label{fig:framework}
\end{figure*}

\subsection{Generative Zero-shot Quantization}
We now describe the optimization procedure of Generative Zero-shot Quantization (GZNQ) with batch normalization and draw the resemblance to the generative modeling.

The feed forward function of a deep neural network at $l$-th layer can be described as:
\begin{align}
F^l=\mathbf{W}^l\phi(\mathbf{W}^{l-1}...\phi(\mathbf{W}^2\phi(\mathbf{W}^1(f_I(z))))\quad\\
\mu_{\mathrm{batch}}=\frac{\sum_{i=1}^N F_i^l}{N}\quad \sigma^2_{\mathrm{batch}}=\frac{\sum_{i=1}^N (F_i^l-\mu_{\mathrm{batch}})^2}{N}
\end{align}
where $\phi(\cdot)$ is an element-wise nonlinearity function and $\mathbf{W}^l$ is the weights of $l$-th layer (freezed during the optimization process). The mean $\mu_{\mathrm{batch}}$ and variance $\sigma^2_{\mathrm{batch}}$ are calculated per-channel over the mini-batches. Furthermore, we denote the input to networks as $f_I(z)$. That is, we have a vector of latent variables $z$ in some high-dimensional space $\mathcal{Z}$ \footnotemark\ but we can easily sample $z_i$ according to $p(z)$. Then, we have a family of deterministic functions $f_I(z_i)$, parameterized by the synthetic samples/images $I$ in pixel space $\mathcal{I}$. $z_i$ determines the parameters such as the number of places by which the pixels of the image are shifted and whether to apply flipping/jitter/shift function $f$ to $I$. Though $I,\mathbf{W}$ are fixed and the mapping $\mathcal{Z}\times\mathcal{I}\rightarrow\mu,\sigma^2$ is deterministic, if $z$ is random, then $\mu_{\mathrm{batch}}$ and $\sigma^2_{\mathrm{batch}}$ are random variables in the space of $\mu$ and $\sigma^2$. We wish to find the optimal $I^*$ such that, even with random flipping/jitter/shift, the computed $\mu_{\mathrm{batch}},\sigma^2_{\mathrm{batch}}$ are still very similar to the Batch-Normalization (BN) statistics \cite{BN} generated by real images.

\footnotetext{Formally, say $z_i$ is a multivariate random variable. The distribution of each of the component random variables can be $\mathrm{Bernoulli}(p)$ or $\mathrm{Unif}(a,b)$.}

Recall Equation (\ref{KL_loss}), minimizing the KL divergence is equivalent to maximizing the following log likelihood
\begin{equation}
\begin{aligned}
I^*&=\arg\max_I\ \text{ln}\ \mathbb{E}_{p(z)}[p(\mu,\sigma^2|z;I)].
\label{eq:loss}
\end{aligned}
\end{equation}
To perform stochastic gradient descent, we need to compute the expectation. However, taking the expectation with respect to $p(z)$ in closed form is not possible in practice. Instead, we take Monte Carlo (MC) estimation by sampling from $p(z)$
\begin{equation}
\begin{aligned}
\mathbb{E}_{p(z)}[p(\mu,\sigma^2|z;I)]\approx\frac{1}{n}\sum_{i=1}^n p(\mu,\sigma^2|z_i;I).
\label{eq:MonteCarlo}
\end{aligned}
\end{equation}
Then, we may approximate the distribution of mean and variance of a mini-batch, \textit{i.e.,} $p(\mu)$ and $p(\sigma^2)$.

\noindent\textbf{Distribution matching}
For the mean variable, we have $\mu_{\mathrm{batch}}=\frac{\sum_{i=1}^N F_i}{N}$ where $F_i$ are features in the sampled mini-batch. We assume that samples of the random variable are i.i.d. then by central limit theorem (CLT) we obtain
\begin{equation}
\mu_{\mathrm{batch}}\sim\mathcal{N}(\mu,\frac{\sigma^2}{N})
\label{eq:distribution}
\end{equation}
for sufficiently large $N$, given $\mu=\mathbb{E}[F]$ and $\sigma^2=\mathbb{E}[(F-\mu_{\mathrm{batch}})^2]$. Similarly, we get
\begin{equation}
\sigma^2_{\mathrm{batch}}\sim\mathcal{N}(\sigma^2,\frac{\mathrm{Var}[(F^l-\mu)^2]}{N}),
\label{eq:distribution}
\end{equation}
where $N$ accounts for the batchsize and $\mathrm{Var}[\cdot]$ is the finite variance, details in Appendix. Then, we further rewrite Equation (\ref{eq:loss}) through (\ref{eq:MonteCarlo}-\ref{eq:distribution}) as follows:
\begin{equation}
\begin{aligned}
\mathcal{L}_{DM}=&\underbrace{\frac{1}{2}\frac{N}{\sigma^2}(\mu_{\mathrm{batch}}-\mu)^2+\frac{1}{2}\frac{N}{\mathrm{Var}[(F^l-\mu)^2]}(\sigma^2_{\mathrm{batch}}-\sigma^2)^2}_{\text{term I}}\\
&+\frac{1}{2}\text{ln }\mathrm{Var}[(F^l-\mu)^2].
\end{aligned}
\label{BN_matching_loss}
\end{equation}
Note that the popular Batch-Normalization matching loss in recent works \cite{ZeroQ, DeepInversion, GDFQ}
\begin{equation*}
\min ||\Vec{\mu}_{\mathrm{batch}}-\Vec{\mu}||_2^2+ ||\Vec{\sigma}^2_{\mathrm{batch}}-\Vec{\sigma}^2||_2^2,
\end{equation*}
which can be regarded as a simplified term I in Eq.(\ref{BN_matching_loss}), leaving out the correlation between $\mu$ and $\sigma^2$ (\textit{i.e.}, the coefficients). Another group of recent methods \cite{DFNQ, KnowledgeWithin} present
\begin{equation}
\min\ \log\frac{\mu_{\mathrm{batch}}}{\mu}-\frac{1}{2}(1-\frac{\sigma^2+(\sigma^2-\sigma^2_{\mathrm{batch}})^2}{\sigma^2_{\mathrm{batch}}}),
\label{eq:kl_mu_var}
\end{equation}
which actually minimizes the following object
\begin{equation*}
\min\ \text{KL}(\mathcal{N}(\mu,\sigma^2)\ ||\ p(F^l)),\quad F^l\sim\mathcal{N}(\mu_\mathrm{batch},\sigma^2_\mathrm{batch}).
\end{equation*}
That is to approximate the distribution $p(F^l)$ defined over features $F^l$ instead of Batch-Normalization statistics $\mu,\sigma^2$ in Eq.(\ref{eq:loss}). Since the parameter space of featuremaps are much larger than $\mu,\sigma^2$, we adopt $\mathcal{L}_{DM}$ to facilitate the learning process.

\noindent\textbf{Pseudo-label generator} Unfortunately, Monte Carlo estimation in (\ref{eq:MonteCarlo}) can be inaccurate given a limited sampling \footnotemark, which may lead to a large gradient variance and poor synthesized results. Hence, it is common practice to introduce regularizations via prior knowledge of $I$, \textit{e.g.}, 

\footnotetext{Consider $I_T=\frac{1}{T}\sum_{t}^Tf(x^t)$, $I_T\rightarrow\mathbb{E}_{x\sim p}[f(x)]$ holds for $T\rightarrow\infty$.}

\begin{equation}
\min\ \mathcal{L}_{CE}(\varphi_{\bm{\omega}}(I),y)
\label{eq:square_loss}
\end{equation}
where $\varphi_{\bm{\omega}}(I)$ produces the categorical probability and $y$ accounts for the ground-truth, which can be regarded as a prior regularization on $I$.

Recently, \cite{hooker2019compressed, CBCM} shows that compressed models are more vulnerable to challenging or complicated examples. Inspired by these findings, we wished to introduce a post-training quantized low-bit network as the pseudo-label generator to help produce ``hard'' samples. However, the selection of bitwidth can be tricky. High-bit networks yield nearly the same distribution as the full-precision counterpart, which results in noisy synthetic results (as pointed out in adversarial attack, high-frequency noise can easily fool the network). Low-bit alternatives fail on easy samples that damage the image diversity. To solve this, we turn to the model ensemble technique, shown in Figure \ref{fig:framework}, then reveal the similarity between (\ref{eq:kl_loss}) and multiple generators/discriminators training in GANs \cite{HoangNLP18, DurugkarGM17}.

\begin{figure}[t]
\begin{minipage}[t]{.235\textwidth}
\centering
\includegraphics[width=0.9\textwidth]{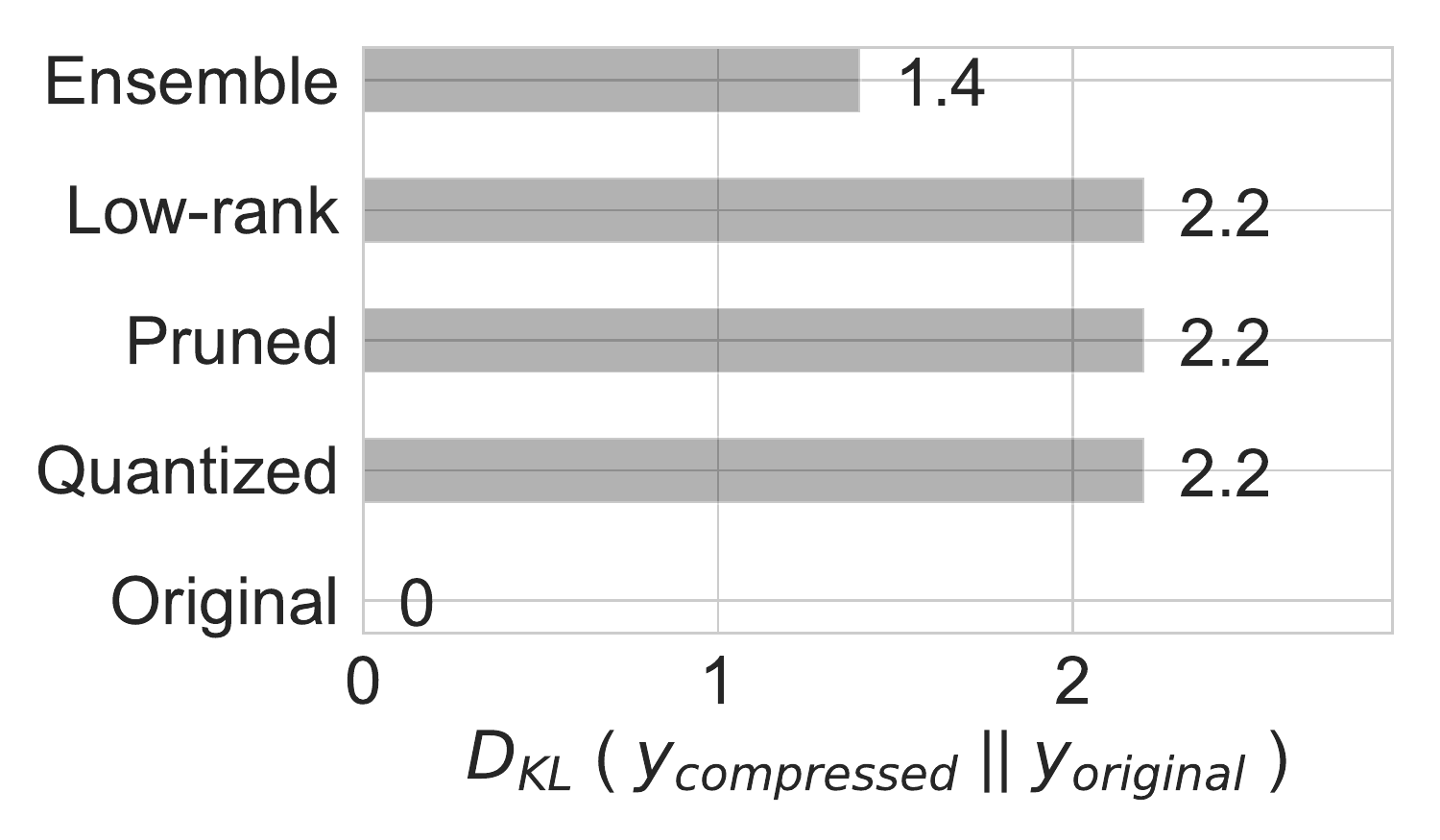}
\\~\\
\includegraphics[width=0.9\textwidth]{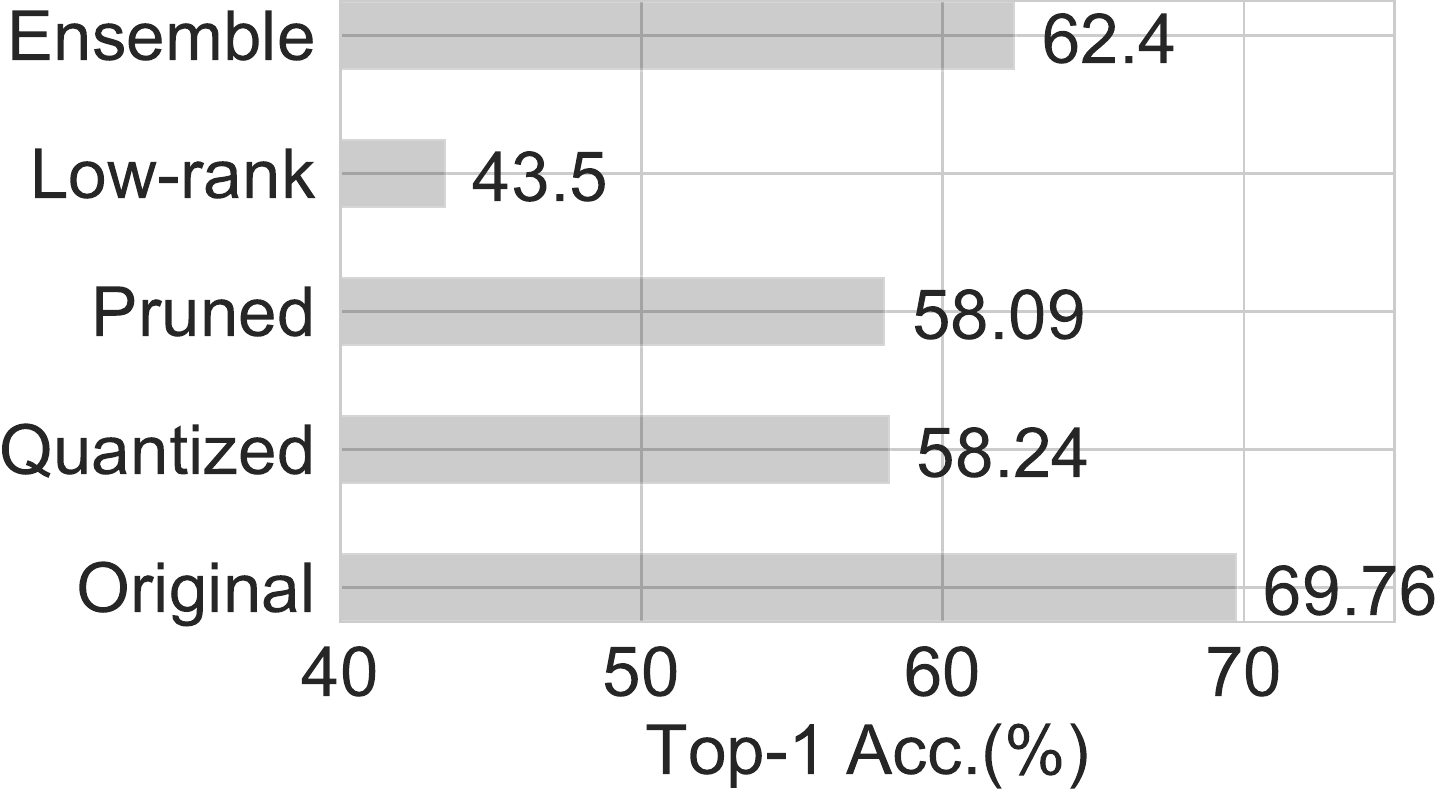}
\subcaption{ResNet-18}
\end{minipage}
\begin{minipage}[t]{.235\textwidth}
\centering
\includegraphics[width=0.9\textwidth]{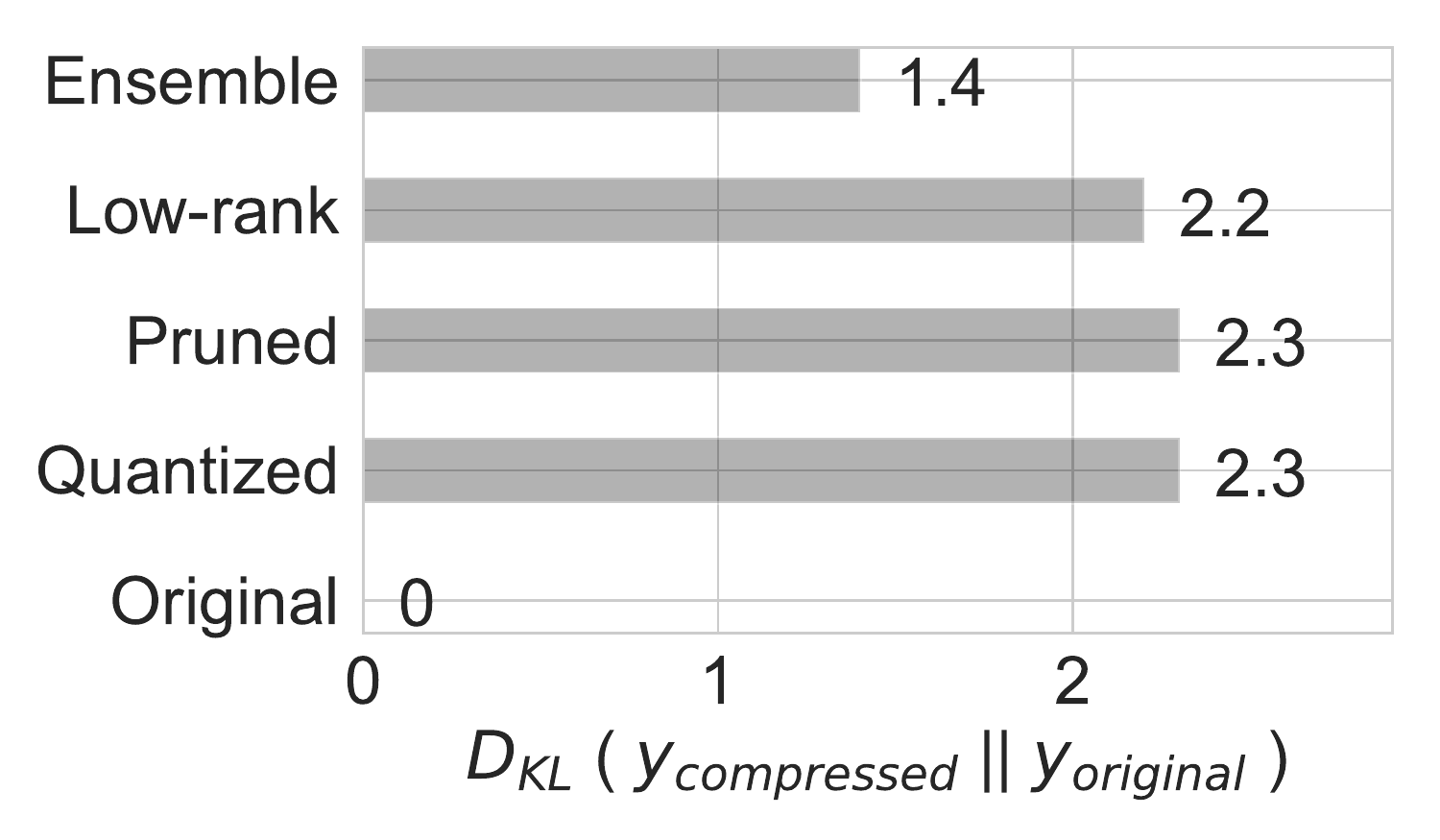}
\\~\\
\includegraphics[width=0.9\textwidth]{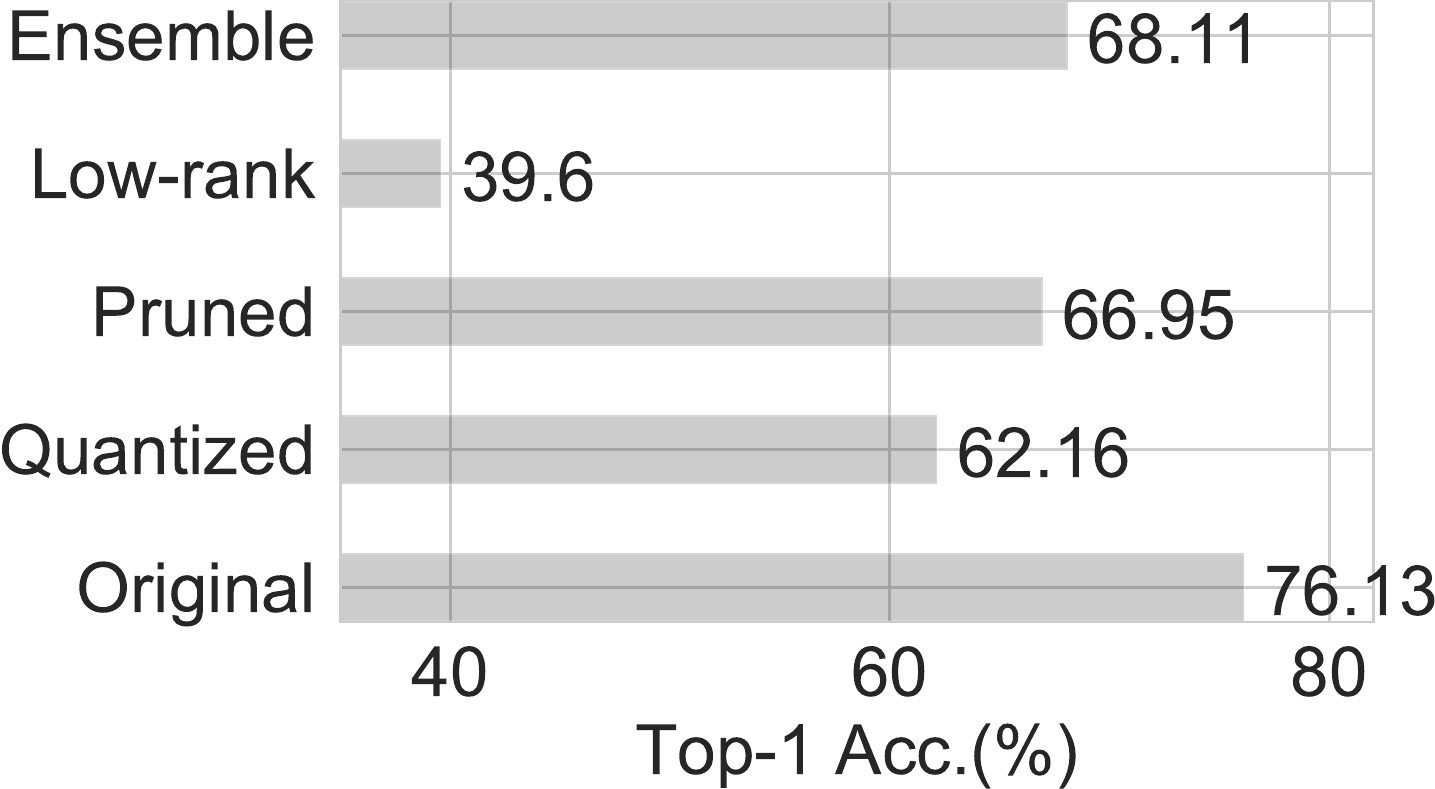}
\subcaption{ResNet-50}
\end{minipage}
\caption{An ensemble of compressed models (same architecture using different post-training compression schemes, \textit{e.g.,} weights MSE quantization + weights magnitude-based unstructured pruning + SVD decomposition of $3\times3$ and FC layers) generates a relatively reliable pseudo-label, which is more similar to the distribution of original network outputs than every single compressed model when the accuracy is comparable.}
\label{fig:ensemble_acc}
\end{figure}

An ensemble of different post-training compressed models generates a similar categorical distribution with the original network (illustrated in Figure \ref{fig:ensemble_acc}) and it is more flexible to adjust the regularization strength than discrete bitwidth selection. Note that ensemble modeling still obtains a small KL distance when the accuracy is relatively low. Here, we get the prior regularization on $I$ as
\begin{equation}
\begin{aligned}
\mathcal{L}_{KL}&=\text{KL}\Big(\varphi_{\bm{\omega}}\left(f_I\left(z\right)\right)|| \frac{1}{M}\sum_{i=1}^M\varphi_{\hat{\bm{\omega}}^i}\left(f_I\left(z\right)\right)\Big)\\
&=\sum_{j=1}^N\varphi_{\bm{\omega}_j}\left(f_{I}\left(z\right)\right)\log\frac{\varphi_{\bm{\omega}_j}\left(f_{I}\left(z\right)\right)}{\frac{1}{M}\sum_{i=1}^M\varphi_{\hat{\bm{\omega}}^i_j}\left(f_{I}\left(z\right)\right))}
\end{aligned}
\label{eq:kl_loss}
\end{equation}
where $\hat{\bm{\omega}}^i$ refers to the $i$-th compressed model. For notational simplicity, we shall in the remaining text denote $\varphi_{\bm{\omega}_j}\left(f_{I}\left(z\right)\right)$ as $\varphi_{\bm{\omega}_j}$. By simply applying the AM-GM inequality to (\ref{eq:kl_loss}), we can easily prove that (\ref{eq:kl_loss}) serves as a ``lower bound'' for the objective with multiple compressed models
\begin{equation}
\begin{aligned}
\sum_{j=1}^N\varphi_{\bm{\omega}_j}&\log\frac{\varphi_{\bm{\omega}_j}}{\frac{1}{M}\sum_{i}\varphi_{\hat{\bm{\omega}}^i_j}}\leq\sum_{j=1}^N\varphi_{\bm{\omega}_j}\log\frac{\varphi_{\bm{\omega}_j}}{(\prod_{i}\varphi_{\hat{\bm{\omega}}^i_j})^{\frac{1}{M}}}\\
&=\frac{1}{M}\sum_{i=1}^M\text{KL}\Big(\varphi_{\bm{\omega}}\left(f_I(z)\right)||\varphi_{\hat{\bm{\omega}}^i}\left(f_I(z)\right)\Big).
\end{aligned}
\label{eq:multi_KL}
\end{equation}
\cite{DurugkarGM17} shows multiple discriminators can alleviate the mode collapse problem in GANs. Here, (\ref{eq:multi_KL}) encourages the synthesized images to be generally ``hard'' for all compressed models (\textit{i.e.,} the large KL divergence corresponds to the disagreement between full-precision networks and compressed models).

\begin{figure}[t]
\centering
\includegraphics[width=0.45\textwidth]{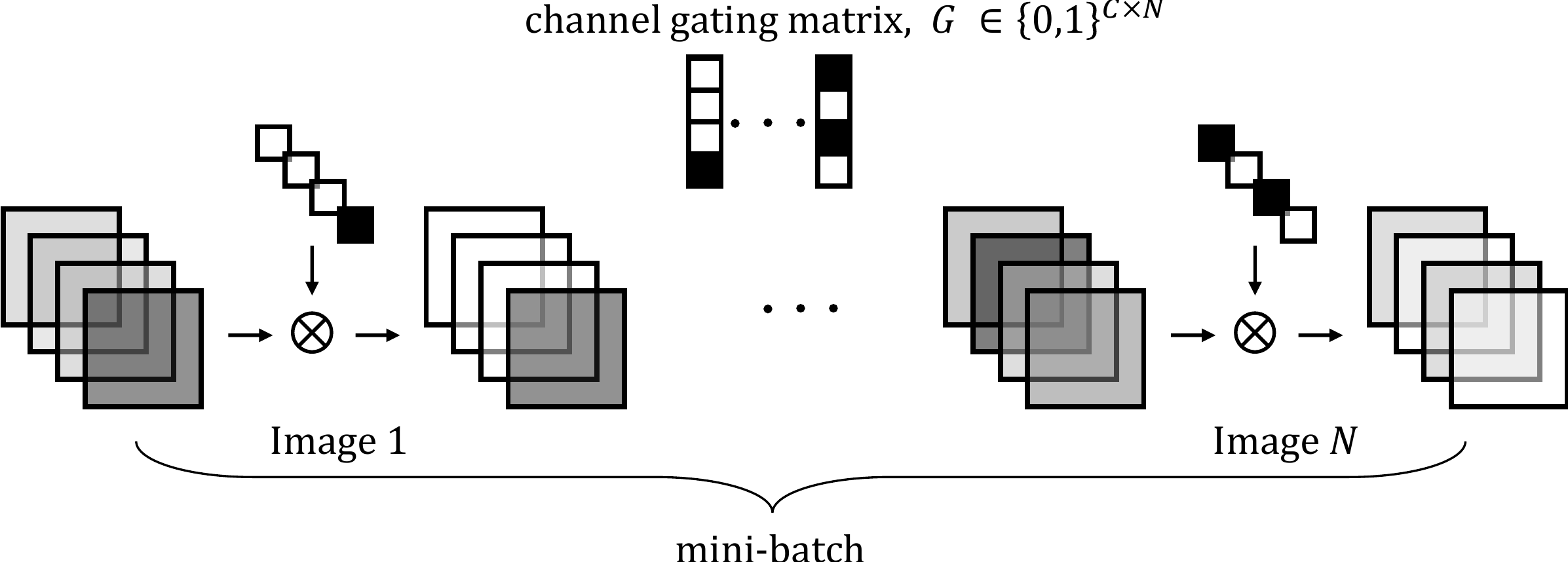}
\caption{Illustration of the channel gating module. Note that each sample has its own binary gating mask $G_i\in\{0,1\}^{1\times c}$. In this case, $c_{hannel}=4$.}
\label{fig:channel_gating}
\end{figure}

\begin{figure}[t]
\begin{minipage}[t]{.13\textwidth}
\centering
\includegraphics[width=\textwidth]{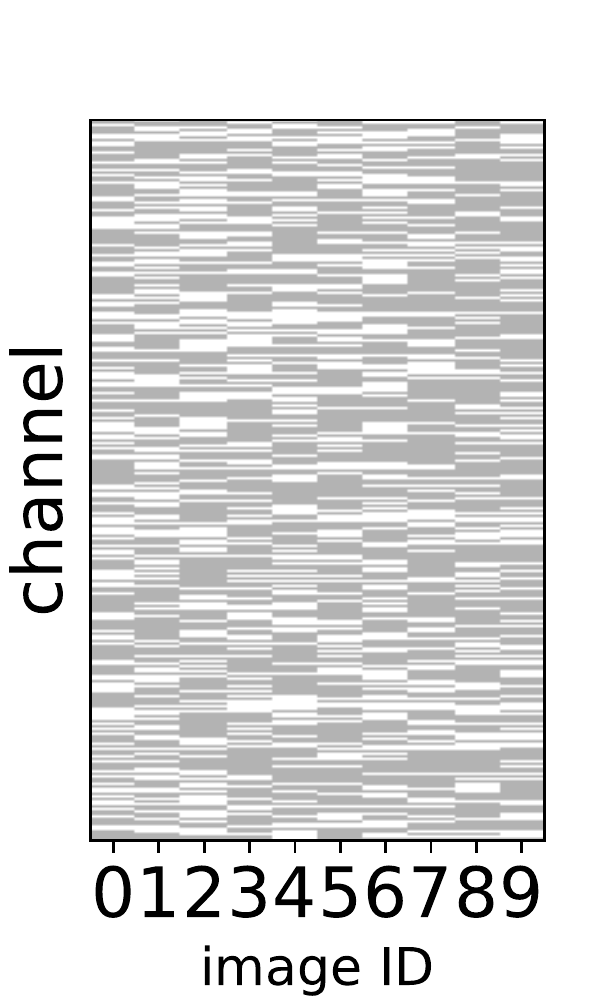}
\subcaption{channel gating}
\end{minipage}
\quad
\begin{minipage}[t]{.3\textwidth}
\centering
\includegraphics[width=\textwidth]{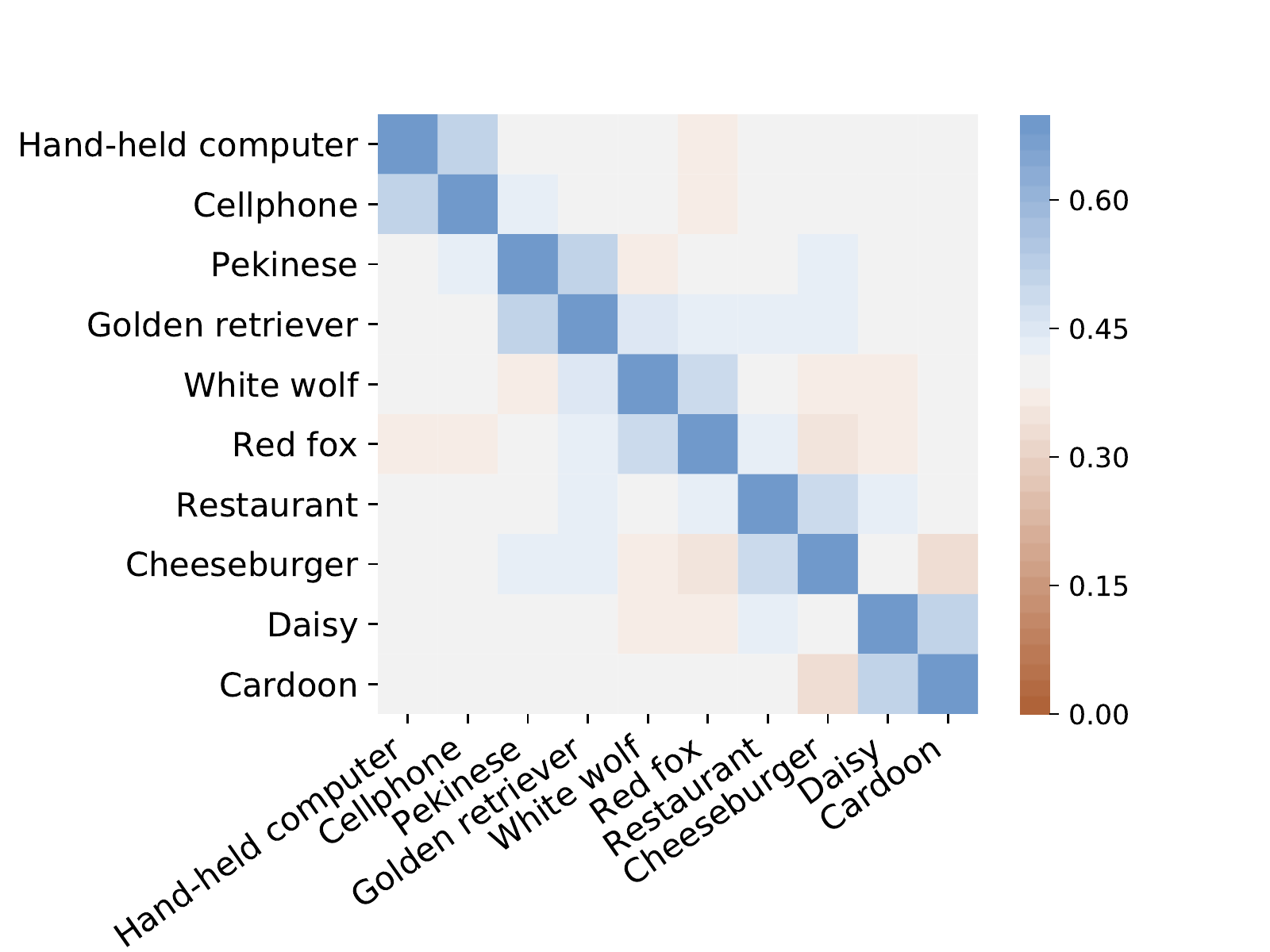}
\subcaption{correlation matrix}
\label{fig:gumble}
\end{minipage}
\caption{We visualize the learned channel gating mask of ten samples in a), generated by ImageNet ResNet-50. There are plenty of zero elements which illustrate the neural redundancy. We further show the correlation matrix of gating masks of ten samples in b). Images belonging to the same main category produce higher responses than irrelevant classes, \textit{e.g.,} the gating mask of cardoon is more similar to daisy than cheeseburger.}
\end{figure}

\noindent\textbf{Channel gating} From the view of connectionism, ``memory'' is created by modifying the strength of the connections between neural units \cite{Smolensky99, sep-connectionism}, \textit{i.e.,} weights matrix. In light of this, we wish to do the optimal surgery \cite{CunDS89, WangS0H18} to enhance the ``memory'' when generating samples belonging to a specific category. More specifically, we use \textit{per-sample} channel pruning to encourage learning more entangled representations, shown in Figure \ref{fig:channel_gating}. Since channel pruning may severely damage the BN statistics, we only apply this setting to the last convolution layer. Fortunately, high-level neurons in deep CNNs are more semantically meaningful, which meets our needs.

We use the common Gumble-Softmax trick \cite{MaddisonMT17} to perform channel gatings
\begin{equation}
G_{i,j}=\left\{ \begin{array}{rcl}
1, & \delta(\frac{\alpha_{i,j}+\log U-\log(1-U)}{\tau})\geq0\\
0, & \text{otherwise}
\end{array}
\right .
\end{equation}
where $\delta$ is the sigmoid function, $U\sim\mathrm{Uniform}(0,1)$ and $\tau$ is the temperature that controls the difference between the softmax and argmax function. Compared with channel pruning \cite{BejnordiBW20} and DARTS \cite{LiuSY19}, we introduce a 2D trainable matrix $\alpha\in\mathbb{R}^{C\times N}$ instead of a vector to better describe each sample/category. Figure \ref{fig:gumble} further illustrates the effect of channel gating that samples of the same main category yield similar masks after training.

\section{Experiments}
We perform experiments on the small-scale CIFAR-10/100 dataset ($32\times32$ pixels) and the complex ImageNet dataset ($224\times224$ pixels, 1k classes). The quantization and fine-tuning process are strictly data-free. We then report the Top-1 accuracy on the validation set.

\subsection{Ablation study}
In this section, we evaluate the effect of each component. All the ablation experiments are conducted on the ImageNet dataset with pre-trained standard ResNet-50 \cite{HeZRS16}. We use the popular Inception Score (IS) \cite{SalimansGZCRCC16} to measure the sample quality despite its notable flaws \cite{IS_note}. As shown in Table \ref{tab:ablation_study}, introducing prior regularization significantly contributes to higher inception scores and other components further improve IS. Since the activations become more semantically meaningful as layers go deeper, we apply (\ref{BN_matching_loss}) to the convolution layers in the last block of ResNets. Table \ref{tab:ablation_study_ensemble} shows ensemble modeling leads to better performance than a single compressed model.

\begin{table*}[t]
\centering
\resizebox{.85\linewidth}{!}{
\begin{tabular}{l c c c c c c c c}
\toprule[1.5pt]
Dataset & Pre-trained Model & Method & W bit & A bit & Quant Acc (\%) & Acc Drop (\%) & Fine-tuning \\
\midrule
\multirow{16}{*}{CIFAR-10} & \multirow{4}{*}{\makecell[c]{ResNet-20\\(0.27M)}} & ZeroQ \cite{ZeroQ} & 4 & 4 & 79.30 & 14.73 & -- \\
& & Ours & 4 & 4 & 89.06 & 4.07 & -- \\
& & GDFQ \cite{GDFQ} & 4 & 4 & 90.25 & 3.78 & \checkmark \\
& & Ours & 4 & 4 & \textbf{91.30} & 1.83 & \checkmark \\
\cline{2-8}
& \multirow{4}{*}{\makecell[c]{ResNet-44\\(0.66M)}} & Knowledge Within \cite{KnowledgeWithin} & 4 & 4 & 89.10 & 4.13 & --\\
& & Ours & 4 & 4 & 91.46 & 2.92 & -- \\
& & Knowledge Within \cite{KnowledgeWithin} & 4$^\dag$ & 8 & 92.25 & 0.99 &-- \\
& & Ours & 4 & 8 & \textbf{93.57} & 0.83 &-- \\
\cline{2-8}
& \multirow{4}{*}{\makecell[c]{WRN16-1\\(0.17M)}} & DFNQ \cite{DFNQ} & 4 & 8 & 88.91 & 2.06 & \checkmark \\
& & Ours & 4 & 4 & \textbf{89.00} & 2.38 & \checkmark \\
& & DFNQ \cite{DFNQ} & 4 & 8 & 86.29 & 4.68 &-- \\
& & Ours & 4 & 4 & 87.59 & 3.79 &-- \\
\cline{2-8}
& \multirow{4}{*}{\makecell[c]{WRN40-2\\(2.24M)}} & DFNQ \cite{DFNQ} & 4 & 8 & 94.22 & 0.55 & \checkmark \\
& & Ours & 4 & 4 & \textbf{94.81} & 0.37 & \checkmark \\
& & DFNQ \cite{DFNQ} & 4 & 8 & 93.14 & 1.63 & -- \\
& & Ours & 4 & 4 & 94.06 & 1.09 & --\\
\midrule\midrule
\multirow{8}{*}{CIFAR-100} & \multirow{4}{*}{\makecell[c]{ResNet-20\\(0.28M)}} & ZeroQ \cite{ZeroQ} & 4 & 4 & 45.20 & 25.13 & -- \\
& & Ours & 4 & 4 & 58.99 & 10.18 & -- \\
& & GDFQ \cite{GDFQ} & 4 & 4 & 63.58 & 6.75 & \checkmark \\
& & Ours & 4 & 4 & \textbf{64.37} & 4.80 & \checkmark \\
\cline{2-8}
& \multirow{4}{*}{\makecell[c]{ResNet-18 \\(11.2M)}} & DFNQ \cite{DFNQ} & 4 & 8 & 75.15 & 2.17 & \checkmark \\
& & Ours & 4 & 5 & \textbf{75.95} & 3.16 & \checkmark \\
& & DFNQ \cite{DFNQ} & 4 & 8 & 71.02 & 6.30 & -- \\
& & Ours & 4 & 5 & 71.15 & 7.96 & -- \\
\bottomrule
\end{tabular}
}
\caption{Results of zero-shot quantization methods on CIFAR-10/100. ``W bit'' means weights quantization bitwidth and ``A bit'' is quantization bits for activations. ``Fine-tuning'' refers to re-training on the generated images using knowledge distillation. $^\dag$ indicates first and last layers are in 8-bit. We directly cite the best results reported in the original zero-shot quantization papers (ZeroQ 4-bit activations from \cite{GDFQ}). }
\label{tab:cifar_acc}
\end{table*}

\subsection{Generative settings}
As shown in Figure \ref{fig:framework}, GZNQ is a two-stage scheme. In this section, we detail the generative settings in the first stage. For CIFAR models, we first train full-precision networks from scratch (initial learning rate 0.1 with cosine scheduler; weight decay is $1e^{-4}$; all networks are trained for 300 epochs by SGD optimizer). Then, we utilize the proposed distribution matching loss, KL divergence loss, channel gating, and CE loss to optimize the synthesized images. More specifically, we use Adam optimizer (beta1 is set to 0.3 and beta2 is 0.9) with a learning rate of 0.4 and generate $32\times32$ images in a mini-batch of 128. The weight of BN matching loss is 0.03 and 0.05 for the KL divergence loss. We first half the image size to speed up training via $2\times2$ average downsampling. After 2k iterations, we use the full resolution images to optimize for another 2k iteration. In this work, we assume $z_i$ to be a three-dimensional random variable such as $z_{i,0}\sim\mathrm{B}(0.5)$ and $z_{i,1}.z_{i,2}\sim\mathrm{U}(-30,30)$, which determines whether to flip images and move images along any dimension by any number of pixels. We follow most settings of CIFAR-10/100 in ImageNet experiments. Since BN statistics and pseudo-labels are dataset-dependent, we adjust the weight of BN matching loss and KL divergence loss to 0.01 and 0.1 respectively. ImageNet pre-trained models are downloaded from torchvision model-zoo directly \cite{PaszkeGMLBCKLGA19}.

In our experiments, we do observe that networks trained only with random $256\times N$/$N\times256$ cropping and flipping contribute to high-quality images but the accuracy is relatively lower than the official model. This finding is consistent with the policy in differential privacy \cite{AbadiCGMMT016}. Since we focus on generative modeling and network quantization, more ablation studies on this part will be our future works.

\begin{table*}[t]
\centering
\resizebox{.85\linewidth}{!}{
\begin{tabular}{c c c c c c c c}
\toprule[1.5pt]
Pre-trained Model & Method & W bit & A bit & Quant Top-1 (\%) & Top-1 Drop (\%) & Real-data & Fine-tuning \\
\midrule
\multirow{7}{*}{\makecell[c]{ResNet-50\\(25.56M)}} & DoReFa \cite{DoReFa} & 4 & 4 & 71.4 & 5.5 & \checkmark & \checkmark \\
& Ours & 4 & 4 & \textbf{72.7} & 3.4 & -- & \checkmark \\
& OMSE \cite{OMSE} & 4 & 32 & 67.4 & 8.6 & -- & -- \\
& Ours & 4 & 6 & 68.1 & 8.1 & -- & -- \\
\cline{2-8}
& OCS \cite{OCS} & 6 & 6 & 74.8 &1.3 & \checkmark & -- \\
& ACIQ \cite{ACIQ} & 6 & 6 & 74.3 &1.3 & \checkmark & -- \\
& Ours & 6 & 6 & \textbf{75.5} & 0.6 & -- & -- \\
\midrule
\multirow{10}{*}{\makecell[c]{ResNet-18\\(11.69M)}} & ZeroQ \cite{ZeroQ} & 4 & 4 & 26.0 & 45.4 & -- & -- \\
& GDFQ \cite{GDFQ} & 4 & 4 & 33.3 & 38.2 & \checkmark & -- \\
& Ours & 4 & 4 & 56.8 & 12.9 & -- & -- \\
& Knowledge Within \cite{KnowledgeWithin} & 4$^\dag$ & 4 & 55.5 & 14.3 & -- & -- \\
& Ours & 4$^\dag$ & 4 & \textbf{58.9} & 10.9 & -- & -- \\
\cline{2-8}
& GDFQ \cite{GDFQ} & 4 & 4 & 60.6 & 10.9 & -- & \checkmark \\
& Ours & 4 & 4 & \textbf{64.5} & 5.30 & -- & \checkmark \\
\cline{2-8}
& Integer-Only \cite{JacobKCZTHAK18} & 6 & 6 & 67.3 & 2.46 & \checkmark & \checkmark \\
& DFQ \cite{DFQ} & 6 & 6 & 66.3 & 3.46 & -- & -- \\
& Ours & 6 & 6 & \textbf{69.0} & 0.78 & -- & -- \\
\midrule
\multirow{7}{*}{\makecell[c]{MobileNetv2\\(3.51M)}}
& Knowledge Within \cite{KnowledgeWithin} & 4$^\ddag$ & 4 & 16.10 & 55.78 & -- & -- \\
& Ours & 4 & 4 & \textbf{53.53} & 17.87 & -- & -- \\
\cline{2-8}
& Integer-Only \cite{JacobKCZTHAK18} & 6 & 6 & 70.90 & 0.85 & \checkmark & \checkmark \\
& Ours & 6 & 6 & \textbf{71.12} & 0.63 & -- & \checkmark \\
\cline{2-8}
& DFQ \cite{DFQ} & 8 & 8 & 71.19 & 0.38 & -- & -- \\
& Knowledge Within \cite{KnowledgeWithin} & 8 & 8 & 71.32 & 0.56 & -- & -- \\
& Ours & 8 & 8 & \textbf{71.38} & 0.36 & -- & -- \\
\bottomrule
\end{tabular}
}
\caption{Quantization results on ImageNet. ``Real-data'' means using original dataset as the calibration set to quantize activations or fine-tune weights. ``Fine-tuning'' refers to re-training with KD on the generated images (if no real data is required) or label-based fine-tuning. $^\dag$ indicates first and last layers are in 8-bit. $^\ddag$ means 8-bit $1\times1$ convolution layer. We directly cite the best results reported in the original papers (ZeroQ from \cite{GDFQ}).}
\label{tab:imagenet_acc}
\end{table*}

\begin{table}[t]
\centering
\resizebox{.99\linewidth}{!}{
\begin{tabular}{c c c c c c}
\toprule[1.5pt]
BN & CE-Loss & Pseudo-label & BN+ & Gating & Inception Score $\uparrow$ \\
\midrule
\checkmark & & & & & 7.4 \\
\checkmark & \checkmark & & & & 43.7 \\
\checkmark &\checkmark & \checkmark & & & 74.0\\
\checkmark & \checkmark & \checkmark & \checkmark & & 80.6\\
\checkmark & \checkmark & \checkmark &\checkmark & \checkmark & \textbf{84.7} \\
\bottomrule
\end{tabular}
}
\caption{Impact of the proposed modules on GZNQ. ``BN+'' refers to Equation (\ref{BN_matching_loss}). Following GAN works \cite{BigGAN, SAGAN}, we use Inception Score (IS) to evaluate the visual fidelity of generated images.}
\label{tab:ablation_study}
\end{table}

\begin{table}[t]
\centering
\resizebox{.99\linewidth}{!}{
\begin{tabular}{l c c c c c}
\toprule[1.5pt]
& Quantization & Pruning & Low-rank & Ensemble (\ref{eq:kl_loss}) & Ensemble (\ref{eq:multi_KL}) \\
\midrule
IS $\uparrow$ & 71.9$\pm$3.85 & 73.0$\pm$2.47 & 81.3$\pm$2.16 & 84.7$\pm$1.76 & 82.7$\pm2.90$ \\
\bottomrule
\end{tabular}
}
\caption{Ablation study on the effect of model ensemble in the Pseudo-label generator. IS stands for Inception Score.}
\label{tab:ablation_study_ensemble}
\end{table}

\begin{table}[t]
\centering
\resizebox{.99\linewidth}{!}{
\begin{tabular}{c c c c c}
\toprule[1.5pt]
\multirow{3}{*}{\makecell[c]{Calibration\\ Dataset}} & \multicolumn{4}{c}{Quantized Model Acc. (\%)}\\
\cline{2-5}
& \makecell[c]{ResNet-20\\(CIFAR-10)} & \makecell[c]{WRN40-2\\(CIFAR-10)} & \makecell[c]{ResNet-20\\(CIFAR-100)} & \makecell[c]{ResNet-18\\(CIFAR-100)} \\
\midrule
SVHN & 24.81 & 50.11 & 5.43 & 7.11 \\
CIFAR-100 & 88.45 & 92.94 & \textbf{60.29} & \textbf{76.63} \\
CIFAR-10 & \textbf{90.06} & \textbf{94.01} & 57.29 & 75.39 \\
Ours & 89.06 & \textbf{94.06} & 58.99 & 71.15 \\
\midrule
FP32 & 93.13 & 95.18& 69.17 & 79.11\\
\bottomrule
\end{tabular}
}
\caption{Impact of using different calibration datasets for 4-bit weights and 4-bit activation post-training quantization. Our synthesized dataset achieves comparable performance to the real images in most cases.}
\label{tab:calibration}
\end{table}

\begin{table}[t]
\centering
\resizebox{.95\linewidth}{!}{
\begin{tabular}{lc c c}
\toprule[1.5pt]
Method & Resolution & GAN & Inception Score $\uparrow$ \\
\midrule
BigGAN-deep~\cite{BigGAN} & $256$ & \checkmark & $202.6$ \\
BigGAN~\cite{BigGAN} & $256$ & \checkmark & $178.0$ \\
SAGAN~\cite{SAGAN} & $128$ & \checkmark & $52.5$ \\
SNGAN~\cite{SNGAN} & $128$ & \checkmark & $35.3$ \\
\midrule
GZNQ & $224$ & - & \textbf{84.7}$\pm$2.8 \\
DeepInversion ~\cite{DeepInversion} & $224$ & - & $60.6$ \\
DeepDream~\cite{DeepDream} & $224$ & - & $6.2$ \\
ZeroQ~\cite{ZeroQ} & $224$ & - & $2.8$ \\
\bottomrule
\end{tabular}
}
\caption{Inception Score (IS, higher is better) of various methods on ImageNet. SNGAN score reported in~\cite{howgoodisgan} and DeepDream score from~\cite{DeepInversion}. The bottom four schemes are data-free and utlizing ImageNet pre-trained ResNet-50 to obtain synthesized images.}
\label{tab:imagenet_is_score}
\end{table}

\subsection{Quantization details}
Low-bit activation quantization typically requires a calibration set to constrain the dynamic range of intermediate layers to a finite set of fixed points. As shown in Table \ref{tab:calibration}, it is crucial to collect images sampled from the same domain as the original dataset. To fully evaluate the effectiveness of GZNQ, we use synthetic images as the calibration set for sampling activations \cite{abs-1806-08342, HubaraCSEB17}, and BN statistics \cite{PBNN, learningCompression}. Besides, we use MSE quantizer \cite{SungSH15, learningCompression} for both weights and activations quantization, though, it is sensitive towards outliers. In all our experiments, floating-point per-kernel scaling factors for the weights and a per-tensor scale for the layer's activation values are considered. We keep a copy of full-precision weights to accumulate gradients then conduct quantization in the forward pass \cite{HubaraCSEB17, KnowledgeWithin}. Additionally, bias terms in convolution and fully-connected layers, gradients, and Batch-Normalization layers are kept in floating points. We argue that advanced calibration methods \cite{UporDown, wang2020unsupervised, GDFQ} or quantization schemes \cite{OMSE, PACT, JacobKCZTHAK18} coupled with GZNQ may further improve the performance.

For our data-free training-aware quantization, \textit{i.e.,} fine-tuning, we follow the setting in \cite{GDFQ, DFNQ} to utilize the vanilla Knowledge Distillation (KD) \cite{HintonVD15} between the original network and the compressed model. Since extra data augmentations can be the game-changer in fine-tuning, we use the common 4 pixels padding with $32\times32$ random cropping for CIFAR and the official PyTorch pre-processing for ImageNet, \textit{without bells and whistles}. The initial learning rate for KD is 0.01 and trained for 300/100 epochs on CIFAR/ImageNet, decayed every $\frac{N}{3}$ iterations with a multiplier of 0.1. The batch size is 128 in all experiments. We also follow \cite{DeepInversion, GDFQ} to fix batch normalization statistics during fine-tuning on ImageNet. All convolution and fully-connected layers are quantized to 4/6-bit, unless specified, including the first and last layer. The synthesized CIFAR dataset consists of 50k images and ImageNet has roughly 100 images per category.

\subsection{Comparisons on benchmarks}
We compare different zero-shot quantization methods \cite{ZeroQ, KnowledgeWithin, DFNQ, GDFQ} on CIFAR-10/100 and show the results in Table \ref{tab:cifar_acc}. Our methods consistently outperform other state-of-the-art approaches in 4-bit settings. Furthermore, benefiting from the high quality generated images, the experimental results in Table \ref{tab:imagenet_acc} illustrate that, as the dataset gets larger, GZNQ still obtains notably improvements over baseline methods. Due to the different full-precision baseline accuracy reported in previous works, we also include the accuracy gap between floating-point networks and quantized networks in our comparisons. We directly cite the results in original papers to make a fair comparison, using the same architecture. In all experiments, we fine-tune the quantized models on the synthetic dataset generated by their corresponding full-precision networks.

\begin{figure}[t]
\begin{minipage}[t]{.075\textwidth}
\centering
\includegraphics[width=0.99\textwidth]{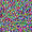}
\includegraphics[width=0.99\textwidth]{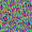}
\includegraphics[width=0.99\textwidth]{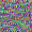}
\subcaption{Noise}
\end{minipage}
\begin{minipage}[t]{.075\textwidth}
\centering
\includegraphics[width=0.99\textwidth]{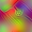}
\includegraphics[width=0.99\textwidth]{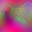}
\includegraphics[width=0.99\textwidth]{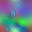}
\subcaption{Deep Dream\cite{DeepDream}}
\end{minipage}
\begin{minipage}[t]{.075\textwidth}
\centering
\includegraphics[width=0.99\textwidth]{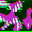}
\includegraphics[width=0.99\textwidth]{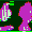}
\includegraphics[width=0.99\textwidth]{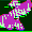}
\subcaption{DAFL\cite{DAFL}}
\end{minipage}
\begin{minipage}[t]{.075\textwidth}
\centering
\includegraphics[width=0.99\textwidth]{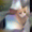}
\includegraphics[width=0.99\textwidth]{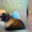}
\includegraphics[width=0.99\textwidth]{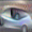}
\subcaption{Deep Inversion\cite{DeepInversion}}
\end{minipage}
\begin{minipage}[t]{.16\textwidth}
\centering
\includegraphics[width=0.47\textwidth]{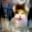}
\includegraphics[width=0.47\textwidth]{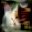}
\includegraphics[width=0.47\textwidth]{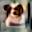}
\includegraphics[width=0.47\textwidth]{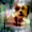}
\includegraphics[width=0.47\textwidth]{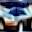}
\includegraphics[width=0.47\textwidth]{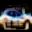}
\subcaption{GZNQ}
\end{minipage}
\caption{Synthetic samples generated by a CIFAR-10 pre-trained ResNet-34 at a $32\times32$ resolution. We directly cite the best visualization results reported in \cite{DeepInversion}.}
\label{fig:cifar_cmp}
\end{figure}

\begin{figure}[t]
\begin{minipage}[t]{.235\textwidth}
\centering
\includegraphics[width=0.99\textwidth]{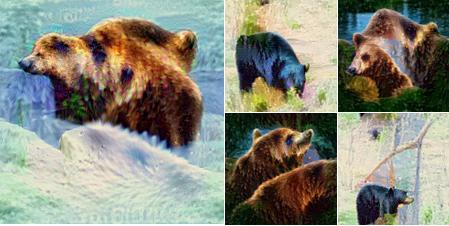}
\end{minipage}
\begin{minipage}[t]{.235\textwidth}
\centering
\includegraphics[width=0.99\textwidth]{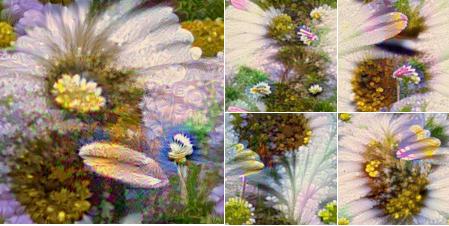}
\end{minipage}

\begin{minipage}[t]{.235\textwidth}
\centering
\includegraphics[width=0.99\textwidth]{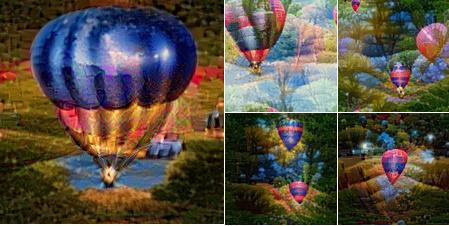}
\end{minipage}
\begin{minipage}[t]{.235\textwidth}
\centering
\includegraphics[width=0.99\textwidth]{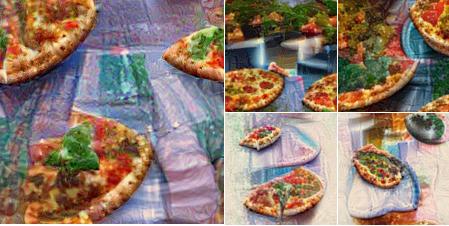}
\end{minipage}

\begin{minipage}[t]{.235\textwidth}
\centering
\includegraphics[width=0.99\textwidth]{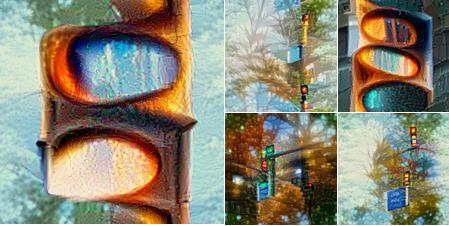}
\end{minipage}
\begin{minipage}[t]{.235\textwidth}
\centering
\includegraphics[width=0.99\textwidth]{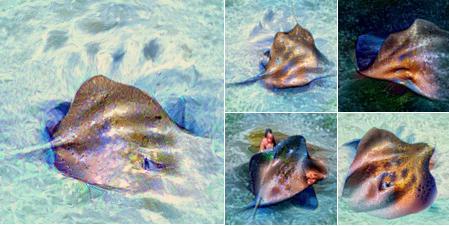}
\end{minipage}

\begin{minipage}[t]{.235\textwidth}
\centering
\includegraphics[width=0.99\textwidth]{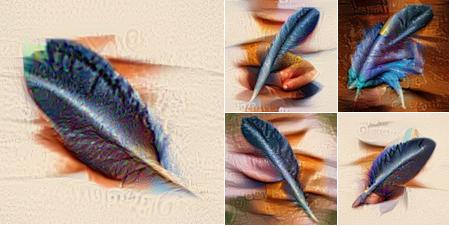}
\end{minipage}
\begin{minipage}[t]{.235\textwidth}
\centering
\includegraphics[width=0.99\textwidth]{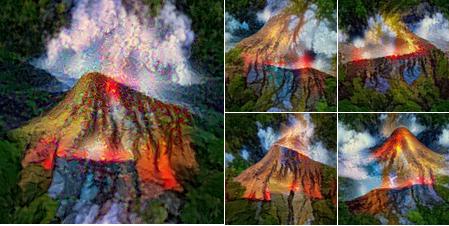}
\end{minipage}
\caption{ImageNet samples generated by our GZNQ ResNet-50 model at $224\times224$ resolution : bear, daisy, balloon, pizza, stoplight, stingray, quill, volcano. }
\label{fig:imagenet}
\end{figure}

\begin{figure}[t]
\begin{minipage}[t]{.09\textwidth}
\centering
\includegraphics[width=0.99\textwidth]{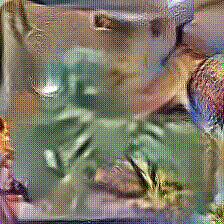}
\includegraphics[width=0.99\textwidth]{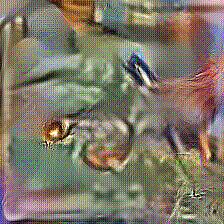}
\includegraphics[width=0.99\textwidth]{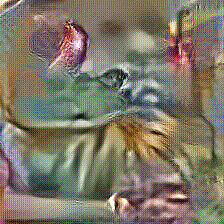}
\includegraphics[width=0.99\textwidth]{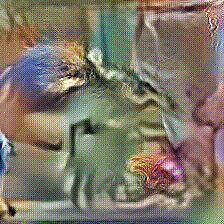}
\subcaption{ZeroQ\cite{ZeroQ}}
\end{minipage}
\begin{minipage}[t]{.09\textwidth}
\centering
\includegraphics[width=0.99\textwidth]{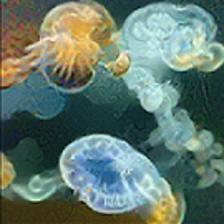}
\includegraphics[width=0.99\textwidth]{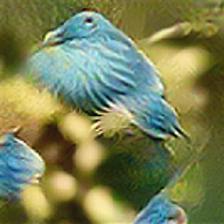}
\includegraphics[width=0.99\textwidth]{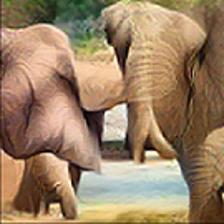}
\includegraphics[width=0.99\textwidth]{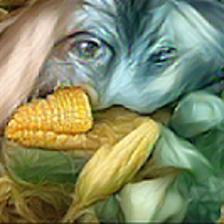}
\subcaption{Knowledge Within\cite{KnowledgeWithin}}
\end{minipage}
\begin{minipage}[t]{.09\textwidth}
\centering
\includegraphics[width=0.99\textwidth]{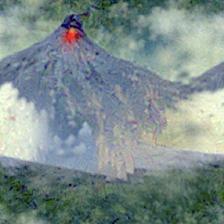}
\includegraphics[width=0.99\textwidth]{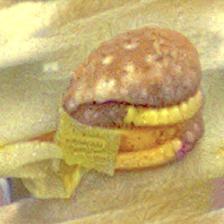}
\includegraphics[width=0.99\textwidth]{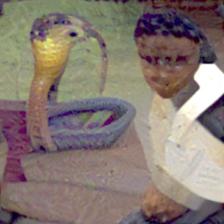}
\includegraphics[width=0.99\textwidth]{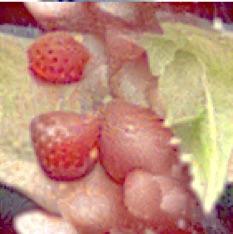}
\subcaption{\\Deep Inversion\cite{DeepInversion}}
\end{minipage}
\begin{minipage}[t]{.1835\textwidth}
\centering
\includegraphics[width=0.486\textwidth]{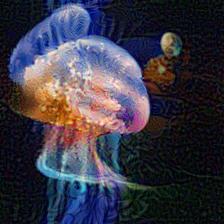}
\includegraphics[width=0.486\textwidth]{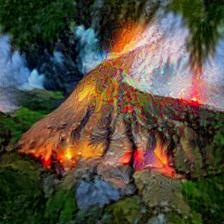}
\includegraphics[width=0.486\textwidth]{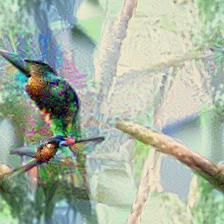}
\includegraphics[width=0.486\textwidth]{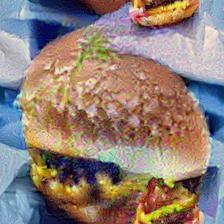}
\includegraphics[width=0.486\textwidth]{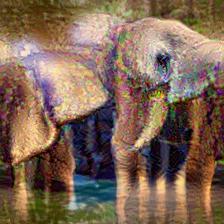}
\includegraphics[width=0.486\textwidth]{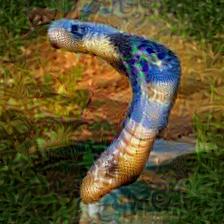}
\includegraphics[width=0.486\textwidth]{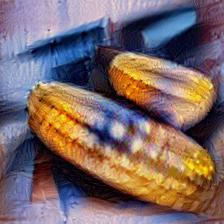}
\includegraphics[width=0.486\textwidth]{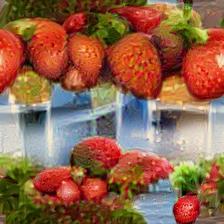}
\subcaption{GZNQ}
\end{minipage}
\caption{Synthetic samples generated by a ImageNet pre-trained ResNet-50 using different methods. We directly cite the best visualization results reported in the original papers.}
\label{fig:imagenet_cmp}
\end{figure}

We further compare our method with \cite{ZeroQ, KnowledgeWithin, DAFL, DeepInversion} on the visual fidelity of images generated on CIFAR-10 and ImageNet. Figure \ref{fig:cifar_cmp}-\ref{fig:imagenet_cmp} show that GZNQ is able to generate images with high fidelity and resolution. We observe that GZNQ images are more realistic than other competitors (appear like cartoon images). Following \cite{DeepInversion}, we conduct quantitative analysis on the image quality via Inception Score (IS) \cite{SalimansGZCRCC16}. Our approach surpasses previous works, that is consistent with visualization results.

\section{Conclusions}
We present generative modeling to describe the image synthesis process in zero-shot quantization. Different from data-driven VAE/GANs that estimates $p(x)$ defined over real images $\mathcal{X}$, we focus on matching the distribution of mean and variance of Batch Normalization layers in the absence of original data. The proposed scheme further interprets the recent Batch Normalization matching loss and leads to high fidelity images. Through extensive experiments, we have shown that GZNQ performs well on the challenging zero-shot quantization task. The generated images also serve as an attempt to visualize what a deep convolutional neural network expects to see in real images.

{\small
\bibliographystyle{ieee_fullname}
\interlinepenalty=10
\bibliography{egbib}
}

\end{document}